\documentclass{article}
\usepackage[preprint]{spconf}
\usepackage{amsmath,graphicx}

\copyrightnotice{\copyright\ IEEE 2019}
\toappear{To appear in {\it IEEE International Conference on Image Processing (ICIP), Sept. 22-25, 2019, Taipei, Taiwan}}
\usepackage {color, graphicx}
\usepackage {subfigure}
\usepackage {tikz}
\usetikzlibrary {patterns,shapes,shadows,decorations.pathreplacing}
\usepackage{amssymb}
\usepackage{algorithm,algpseudocode}
\usepackage{booktabs} 
\usepackage{tabularx}
\usepackage{amsmath}
\usepackage{upgreek}
\usepackage{url}
\usepackage{amsmath,amsbsy}

\definecolor{Lightgray}{RGB}{235,235,235}

\newcommand{\etal}{\emph{et~al.}}

\usepackage{url}
\usepackage{hyperref}

\usepackage{xcolor}
\usepackage{amssymb}
\usepackage{pifont}
\newcommand{\cmark}{\ding{51}}%
\newcommand{\xmark}{\ding{55}}%

\tikzset{
  annotated cuboid/.pic={
    \tikzset{%
      every edge quotes/.append style={midway, auto},
      /cuboid/.cd,
      #1
    }
\draw [every edge/.append style={pic actions, dotted, opacity=0.0}, pic actions]
    (0,0,0) coordinate (o) -- ++(-\cubescale*\cubex,0,0) coordinate (a) -- ++(0,-\cubescale*\cubey,0) coordinate (b) edge coordinate [pos=1] (g) ++(0,0,-\cubescale*\cubez)  -- ++(\cubescale*\cubex,0,0) coordinate (c) -- cycle
    (o) -- ++(0,0,-\cubescale*\cubez) coordinate (d) -- ++(0,-\cubescale*\cubey,0) coordinate (e) edge (g) -- (c) -- cycle
    (o) -- (a) -- ++(0,0,-\cubescale*\cubez) coordinate (f) edge (g) -- (d) -- cycle;
  },
  /cuboid/.search also={/tikz},
  /cuboid/.cd,
  width/.store in=\cubex,
  height/.store in=\cubey,
  depth/.store in=\cubez,
  units/.store in=\cubeunits,
  scale/.store in=\cubescale,
  width=10,
  height=10,
  depth=10,
  units=cm,
  scale=.1,
}

\begin {document}

\title {A Two-Stream Siamese Neural Network for Vehicle\\ Re-Identification by Using Non-Overlapping Cameras}

\name{Icaro O. de Oliveira, Keiko V. O. Fonseca and Rodrigo Minetto \thanks {Icaro O. de Oliveira, Keiko V. O. Fonseca and Rodrigo Minetto are with the Graduate Program in Electrical and Computer Engineering (CPGEI) and PPGCA. E-mail: icarofua@gmail.com}}
\address{Federal University of Technology - Paran\'{a} (UTFPR), Curitiba, Brazil}

\maketitle

\begin {abstract}
We describe in this paper a Two-Stream Siamese Neural Network for vehicle re-identification. The proposed network is fed simultaneously with small coarse patches of the vehicle shape's, with $96 \times 96$ pixels, in one stream, and fine features extracted from license plate patches, easily readable by humans, with $96 \times 48$ pixels, in the other one. Then, we combined the strengths of both streams by merging the Siamese distance descriptors with a sequence of fully connected layers, as an attempt to tackle a major problem in the field, false alarms caused by a huge number of car design and models with nearly the same appearance or by similar license plate strings. In our experiments, with 2 hours of videos containing 2982 vehicles, extracted from two low-cost cameras in the same roadway, 546 ft away, we achieved a $F$-measure and accuracy of 92.6\% and 98.7\%, respectively. 
We show that our network, available at {\footnotesize \url{https://github.com/icarofua/siamese-two-stream}}, outperforms other One-Stream architectures, even if they use higher resolution image features.

\end{abstract}

\begin {keywords}
Vehicle Re-identification; Siamese Neural Networks; Vehicle Matching; Travel Time Estimation.
\end {keywords}

\section{Introduction}~\label{cha:intro}

\begin{figure}[!htb]
   \begin{tikzpicture}
     \draw(0.0,1.7) node[] (image) {
       \frame{\includegraphics[width=6.0cm]{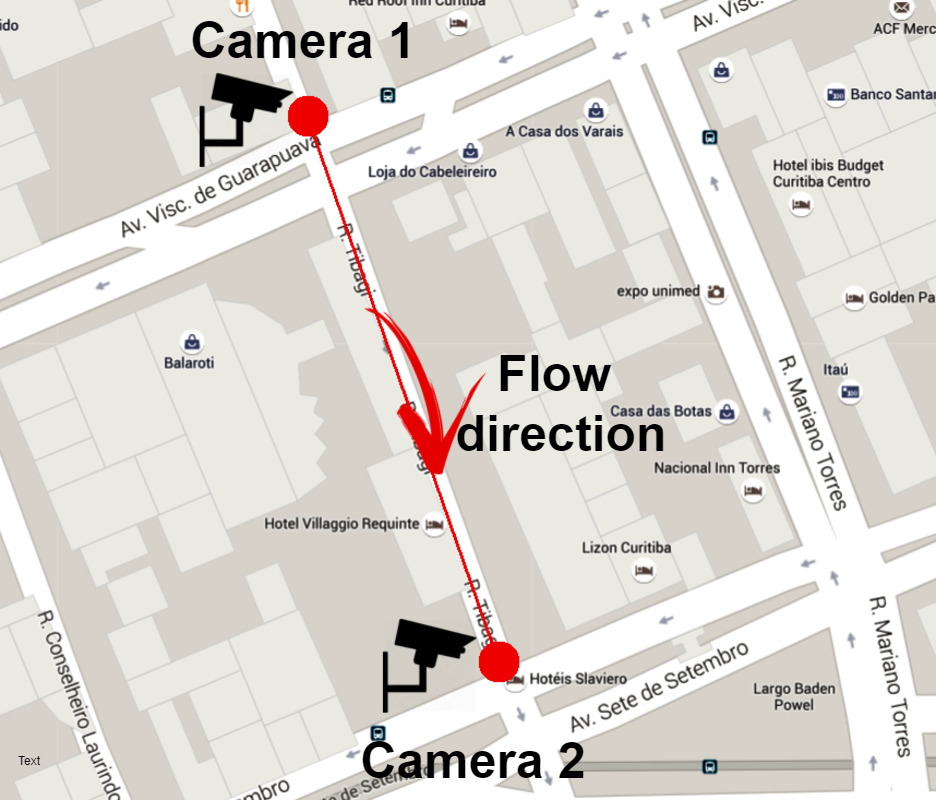}}   
     };
     \draw(3.3,3.2) node[text centered, text width=2.7cm, inner sep=0pt] (image) { \frame{\includegraphics[width=3.5cm]{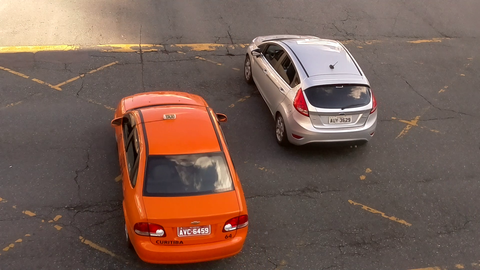}} };
     \draw(3.3,0.2) node[text centered, text width=2.7cm, inner sep=0pt] (image) { \frame{\includegraphics[width=3.5cm]{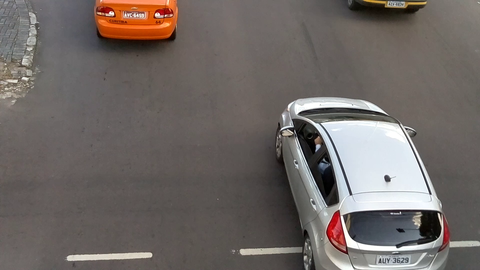}} };
   \end{tikzpicture}
  \caption{System setup: a traffic engineering company placed in two different semaphores a pair of low-cost full-HD cameras properly calibrated and time synchronized. In general, not every vehicle seen in one video appears in the other video.
  }  \label{fig:setup}
\end{figure}

This paper address the problem of matching moving vehicles that appear in two videos taken by multiple cameras with non-overlapping fields of view. See Fig.~\ref{fig:setup}. This is a common sub-problem for several applications in intelligent transportation systems, such as enforcement of road speed limits, criminal investigations, monitoring of commercial transportation vehicles, and traffic management.

Some of these applications traditionally use physical sensors placed over, near, or under the road, such as pressure-sensitive cables and inductive loop detectors~\cite{5763781,5659904}. However, these detectors present limitations such as the same vehicles entering or leaving the road between the two cameras.  Other applications use optical character recognition (OCR) algorithms~\cite{hiercnn} to translate the license plate image regions into character codes, such as ASCII. However, this translation is not straightforward when two or more lanes are recorded at the same time, producing small license plate regions that are very hard to read. Recognition of vehicles by shape and color is not sufficiently reliable either, since vehicles of the same brand and model often look exactly the same \cite{she2004vehicle}.  

For such reasons, in our solution we have opted to identify vehicles across non-overlapping cameras by using an hybrid strategy, that is, we developed a Two-Stream Siamese Neural Network that is fed, simultaneously, with two of the most distinctive and persistent features available, the vehicle's shape and the registration license plate. Then, for fusion of the Two-Streams we concatenate the distance descriptors extracted from each single Siamese network and add fully connected layers for classification. We also show that the combination of small image patches produces a fast network that outperforms other complex architectures, even if they use higher resolution image patches. The rest of this paper is organized as follows.  In Sec.~\ref{sec:related}, we discuss the related work.  In Sec.~\ref{sec:method}, we describe the Two-Stream Siamese Network. Experiments are reported in Sec.~\ref{sec:experiments}. Finally, in Sec.~\ref{sec:conclusions} we state the conclusions.


\section{Related work}~\label{sec:related}




Vehicle re-identification is an active field of research with many algorithms and extensive bibliography~\cite{5763781,5659904,zhong2019poses,8296310,8265213,8036238,tang2017multi}.
The survey of Tian~{\etal}~\cite{6875912} listed this problem as an open challenge for intelligent transportation systems. Traditionally, algorithms for this task were based on the comparison of electromagnetic signatures. However, as observed by Ndoye~\etal~\cite{5659904}, such signature-matching algorithms are exceedingly complex and depends on extensive calibrations or complicated data models.

Video-based algorithms have been proven to be powerful for vehicle re-identification~\cite{5659904,8265213,8036238,tang2017multi,8451776}. Such algorithms need to address \emph{fine-grained vehicle recognition} issues~\cite{7350898}, that is,
to distinguish between subordinate categories with similar visual appearance, caused by a huge number of car design and models with similar appearance. As an attempt to solve these issues many authors proposed to use handed-crafted image descriptors such as SIFT~\cite{Zhang2016}. Recently, inspired by the tremendous progress of the Siamese Neural Networks  Tang~\etal~\cite{tang2017multi}, in 2017, proposed for vehicle re-identification in traffic surveillance environment to fuse deep and hand-crafted features by using a Siamese Triplet Network~\cite{10.1007/978-3-319-24261-3_7}. In 2018, Yan~\etal~\cite{8265213} proposed a novel deep learning metric, a Triplet Loss Function, that takes into account the inter-class similarity and intra-class variance in vehicle models considering only the vehicle's shape. Also in 2018, Liu~\etal~\cite{8036238} proposed a coarse-to-fine vehicle re-identification algorithm that initially filters out the potential matchings by using hand-crafted and deep features based on shape and color and, then they used the license plates in a Siamese Network and a Spatiotemporal re-ranking to refine the search. 

The idea of a two stream convolutional neural networks (CNN) is not new. Ye~\etal~\cite{Ye:2015:ETC:2671188.2749406} proposed an architecture that uses static video frames as input in one stream and optical flow features in the other stream for video classification. Chung~\etal~\cite{chung2017two} also proposed a two-stream architecture composed by two Siamese CNN fed with spatial and temporal information extracted from RGB frames and optical flow vectors for person re-identification. Zagoruyko~\etal~\cite{zagoruyko2015} described distinct Siamese architectures to compare learning image patches. In special, the Central-Surround Two-Stream architecture is similar to the one proposed here.

Finally, some authors~\cite{5763781} use self-adaptive time-window constraints to define upper and lower bounds in order to predict the search space and narrow-down the potential matches. That is, they predict a time-window size based on the camera's distance and the traffic conditions, e.g. free flow or congested. However, we are not trying to solve the travel time estimation problem here, thus, we considered the maximum number of true or false matchings available in order to evaluate the robustness of the architectures.

\section{Two-Stream Siamese Network}~\label{sec:method}

\begin{figure}[!htb]
\begin{tikzpicture}
\tikzset{blockS/.style={draw, rectangle, text centered, drop shadow, fill=white, text width=1.0cm}}
\tikzset{blockL/.style={draw, rectangle, text centered, drop shadow, fill=white, text width=3.2cm}}
\tikzset{blockG/.style={draw, rectangle, text centered, drop shadow, fill=white, text width=3.6cm}}

\path[->](1.0,6.6) node[] {\textbf{\large Camera 1}};
\path[->](5.5,6.6) node[] {\textbf{\large Camera 2}};

\draw(0.0,4.9) node[text centered, text width=2.0cm] (img1) {
{\small Shape \\ 96$\times$96 pixels}\\
\vspace{0.1cm}
\includegraphics[width=2.0cm,height=2.0cm]{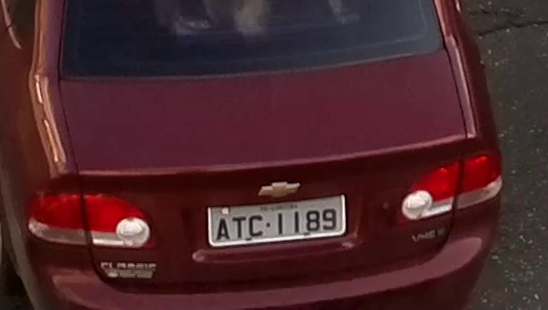}
};

\draw(2.2,4.9) node[text centered, text width=2.2cm] (img2) {
{\small Plate \\ 96$\times$48 pixels}\\
\vspace{0.1cm}
\includegraphics[width=2.0cm,height=1.0cm]{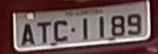}
};

\draw(4.4,4.9) node[text centered, text width=2.2cm] (img3) {
{\small Shape \\ 96$\times$96 pixels}\\
\vspace{0.1cm}
\includegraphics[width=2.0cm,height=2.0cm]{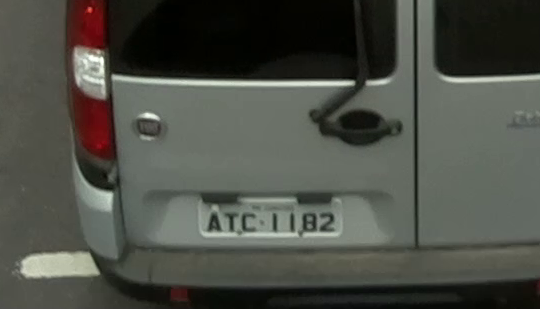}
};

\draw(6.6,4.9) node[text centered, text width=2.2cm] (img4) {
{\small Plate \\ 96$\times$48 pixels}\\
\vspace{0.1cm}
\includegraphics[width=2.0cm,height=1.0cm]{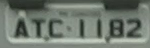}
};


\path[->](0.0,2.4) node[blockS] (cnn1) {
\textbf{CNN}
};

\path[->](2.2,2.4) node[blockS] (cnn2) {
\textbf{CNN}
};

\path[->](1.1,1.3) node[blockL] (d1) {
\textbf{Distance ($\mathbb{L}_1$)}
};

\draw [->] (cnn1) to [out=270,in=110] (d1);
\draw [->] (cnn2) to [out=270,in=70] (d1);

\path[->](4.4,2.40) node[blockS] (cnn3) {
\textbf{CNN}
};

\path[->](6.6,2.40) node[blockS] (cnn4) {
\textbf{CNN}
};

\path[->](5.5,1.3) node[blockL] (d2) {
\textbf{Distance ($\mathbb{L}_1$)}
};

\draw [->] (cnn3) to [out=270,in=110] (d2);
\draw [->] (cnn4) to [out=270,in=70] (d2);

\path[<->] (cnn1) edge[line width=0.3mm] node[fill=white, anchor=center, pos=0.5,font=\bfseries, inner sep=0pt] {W} (cnn2);

\path[<->] (cnn3) edge[line width=0.3mm] node[fill=white, anchor=center, pos=0.5,font=\bfseries, inner sep=0pt] {W} (cnn4);

\path[->](3.3,0.1) node[blockG] (fusion) {
\textbf{Concatenate (Fusion)}
};

\draw [->] (d1) to [out=270,in=130] (fusion);
\draw [->] (d2) to [out=270,in=50] (fusion);

\path[->](3.3,-0.7) node[blockG] (fc1) {
\textbf{Fully Connected (1024)}
};

\path[->](3.3,-1.5) node[blockG] (fc2) {
\textbf{Fully Connected (512)}
};

\path[->](3.3,-2.3) node[blockG] (fc3) {
\textbf{Fully Connected (256)}
};

\path[->](3.3,-3.1) node[blockG] (fc4) {
\textbf{Fully Connected (2)}
};

\draw [->] (fusion) to [] (fc1);
\draw [->] (fc1) to [] (fc2);
\draw [->] (fc2) to [] (fc3);
\draw [->] (fc3) to [] (fc4);

\draw [rounded corners=0.5cm, dashed] (3.5,0.8) rectangle (7.5,2.9) node [midway]{};

\path[->](0.2,0.6) node[] {\textbf{Stream 1}};

\draw [rounded corners=0.5cm, dashed] (-0.9,0.8) rectangle (3.0,2.9) node [midway]{};

\path[->](6.4,0.6) node[] {\textbf{Stream 2}};


\draw [->] (img1) to [out=270,in=90] (cnn1);
\draw [->] (img3) to [out=270,in=90] (cnn2);
\draw [->] (img2) to [out=270,in=90] (cnn3);
\draw [->] (img4) to [out=270,in=90] (cnn4);

\path[->](1.5,-4.2) node[] (out1) {\textbf{Matching}};

\path[->](5.2,-4.2) node[] (out2) {\textbf{\textcolor{red}{Non-Matching}}};

\draw [->] (fc4) to [out=210,in=90] (out1);
\draw [->,thick,red] (fc4) to [out=330,in=90] (out2);

\end{tikzpicture}
\caption{Inference flowchart of the proposed Two-Stream Siamese for Vehicle Matching.}
\label{fig:system-overview}
\end{figure}

The inference flowchart of the proposed Two-Stream Siamese Network is shown in Fig.~\ref{fig:system-overview}. The left stream processes the vehicle's shapes while the right stream the license plates. The network weights $W$ are shared only within each stream. We merged the distance vectors of each Siamese --- whose similarity is measured by a Mahalanobis distance --- and combined the strengths of both features by using a sequence of
fully connected layers with dropout regularization (20\%) in order to avoid over-fitting. Then, we used a softmax activation function to classify matching pairs from non-matching pairs.

\begin{figure}[!htb]
\begin{tikzpicture}[scale=1.0]
\begin{scope}[scale=1.2]

  \pic [fill=white, draw=black] at (0,0) {annotated cuboid={width=1, height=20, depth=16, units=mm}};

  \pic [fill=darkgray!70, draw=black] at (0.6,0) {annotated cuboid={width=2, height=20, depth=16, units=mm}};
  \pic [fill=blue!15, draw=black] at (0.9,-0.24) {annotated cuboid={width=2, height=17, depth=14, units=mm}};

  \pic [fill=darkgray!70, draw=black] at (1.6,-0.24) {annotated cuboid={width=3, height=17, depth=14, units=mm}};
  \pic [fill=blue!15, draw=black] at (2.0,-0.48) {annotated cuboid={width=3, height=14, depth=14, units=mm}};

  \pic [fill=darkgray!70, draw=black] at (2.7,-0.48) {annotated cuboid={width=3, height=14, depth=14, units=mm}};
  \pic [fill=blue!15, draw=black] at (3.1,-0.72) {annotated cuboid={width=3, height=11, depth=14, units=mm}};

  \pic [fill=darkgray!70, draw=black] at (3.9,-0.72) {annotated cuboid={width=4, height=11, depth=14, units=mm}};
  \pic [fill=blue!15, draw=black] at (4.4,-0.96) {annotated cuboid={width=4, height=8, depth=14, units=mm}};

  \pic [fill=darkgray!70, draw=black] at (5.1,-0.96) {annotated cuboid={width=5, height=8, depth=14, units=mm}};
  \pic [fill=blue!15, draw=black] at (5.8,-1.20) {annotated cuboid={width=5, height=5, depth=14, units=mm}};

  \pic [fill=red, draw=black] at (6.8,-1.44) {annotated cuboid={width=8, height=2, depth=2, units=mm}};

{\scriptsize
\draw [decoration={brace,mirror},decorate] (-0.2,-1.70) -- (0.30,-1.70) node [pos=0.5,anchor=north,yshift=-0.05cm] {W$\times$H$\times$3};

\draw [decoration={brace},decorate] (0.7,0.62) -- (1.40,0.62) node [pos=0.5,anchor=south,yshift=0.05cm] {64 filters};
\draw [decoration={brace},decorate] (1.6,0.3) -- (2.6,0.3) node [pos=0.5,anchor=south,yshift=0.05cm] {128 filters};
\draw [decoration={brace},decorate] (2.7,0.05) -- (3.7,0.05) node [pos=0.5,anchor=south,yshift=0.05cm] {128 filters};
\draw [decoration={brace},decorate] (3.9,-0.2) -- (4.9,-0.2) node [pos=0.5,anchor=south,yshift=0.05cm] {256 filters};
\draw [decoration={brace},decorate] (5.1,-0.45) -- (6.4,-0.45) node [pos=0.5,anchor=south,yshift=0.05cm] {512 filters};

\path[->](0.2,0.6) node[] {Input};

\path[->](6.4,-1.8) node[] {FC (512 units)};

}
\end{scope}

\end{tikzpicture}
\caption{Small-VGG: a VGG-based convolutional neural network used in the Two-Stream Siamese Network. The dark gray boxes denote convolutions by using filter kernel size of $3 \times 3$; the light blue boxes denote $2 \times 2$ max-pooling layers; and, the red box depicts a fully connected layer.}
\label{fig:CNN}
\end{figure}
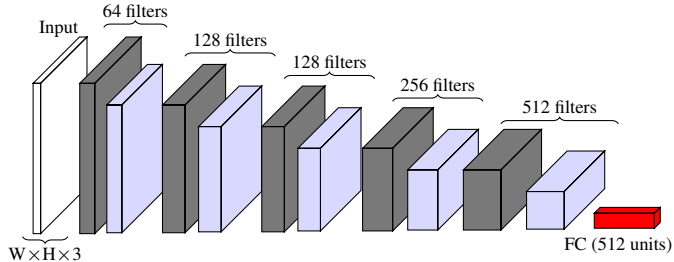

We extracted the vehicle rear end and the vehicle license plate by using the real-time motion detector and algorithms described by Luvizon~\etal~\cite{Luvizon:2016,Minetto:2013}. The CNN used in our Siamese is shown in Fig.~\ref{fig:CNN}. Basically, it is a simplified VGG~\cite{vgg} based network, with a reduced number of layers so as to save computational effort. Each CNN provided a vector with 512 features. Each Distance ($\mathbb{L}_1$) provided a vector with 512 distances. Finally, Concatenate (Fusion) provided a vector with 1024 distances.

\section{Experiments}~\label{sec:experiments}


For our tests, we used 10 videos --- 5 from Camera
1 and 5 from Camera 2 (20 minutes of duration each one) --- recorded
with frame resolution of $1920\times1080$ pixels, at $30.15$ frames per second.
They are summarized in Table~\ref{tab:dataset}.

\begin{table}[!htpb]
   \setlength{\tabcolsep}{1.9pt}
   \def\mc{\multicolumn{2}{c||}}
   \def\df{\normalsize}
   \def\de{\normalsize}

   \centering
   \caption{Dataset information: number of
 vehicles and number of vehicles with a visible license plate in Camera 1 and 2; number of vehicles
 matchings between Camera 1 and 2. }
   \begin{tabular}{||c||c|c||c|c||c||}
         \hline
           \de Set    &  \mc{\de Camera 1} & \mc{\de Camera 2} & \de No.Match.  \\ \hline
                  &  \df \#Vehicles & \df \#Plates & \df \#Vehicles & \df \#Plates &  \\ \hline
           \de 01 & \de 389 & \de   343 & \de     280 &  \de     245 & \de    199  \\
           \de 02 & \de    350 & \de   310 & \de     244 &  \de     227 & \de    174 \\
           \de 03 & \de    340 & \de   301 & \de     274 &  \de     248 & \de    197 \\
           \de 04 & \de    280 & \de   251 & \de     233 &  \de     196 & \de    140 \\
           \de 05 & \de    345 & \de   295 & \de     247 &  \de     194 & \de    159  \\ \hline
        \de Total & \de    1704 & \de   1500 & \de     1278 &  \de     1110 & \de    869 \\ \hline
   \end{tabular}
   \label{tab:dataset}
 \end{table}



\begin{figure}[!htb]
\centering
\begin{tikzpicture}


\draw(0.0,0.0) node[text centered, text width=2.2cm, inner sep=0pt] (cruz1-img1) {
\includegraphics[width=1.3cm,height=1.3cm]{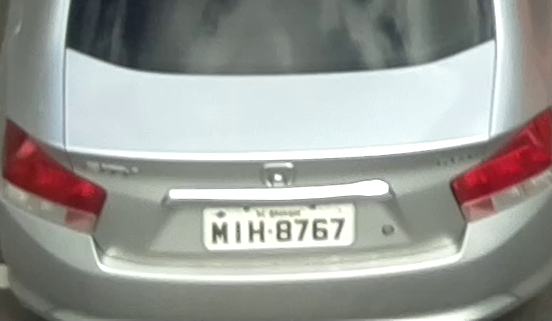}
};

\draw(1.6,0.0) node[text centered, text width=2.2cm, inner sep=0pt] (cruz1-img2) {
\includegraphics[width=1.3cm,height=1.3cm]{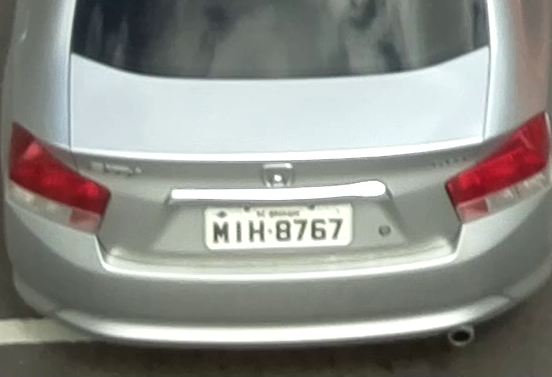}
};

\draw(4.5,0.0) node[text centered, text width=2.2cm, inner sep=0pt] (cruz1-img6) {
\includegraphics[width=1.3cm,height=1.3cm]{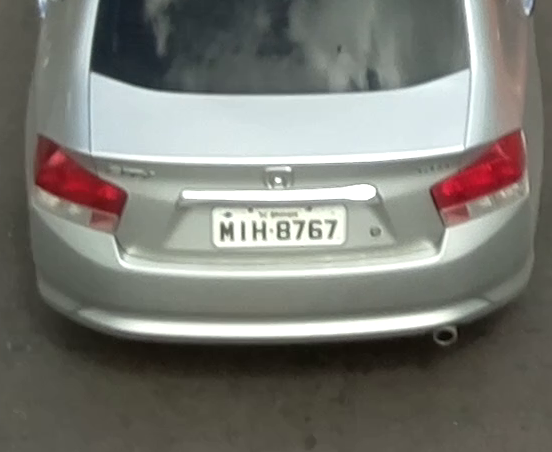}
};

\draw(0.0,2.4) node[text centered, text width=2.2cm, inner sep=0pt] (cruz2-img1) {
\includegraphics[width=1.3cm,height=1.3cm]{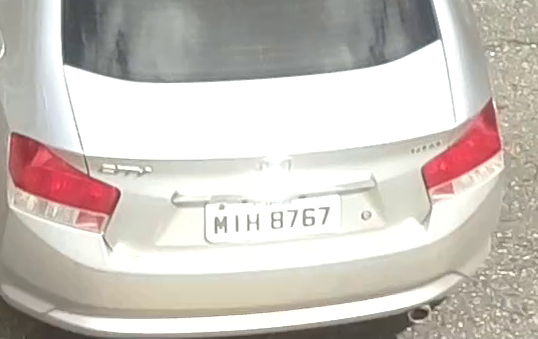}
};

\draw(1.6,2.4) node[text centered, text width=2.2cm, inner sep=0pt] (cruz2-img2) {
\includegraphics[width=1.3cm,height=1.3cm]{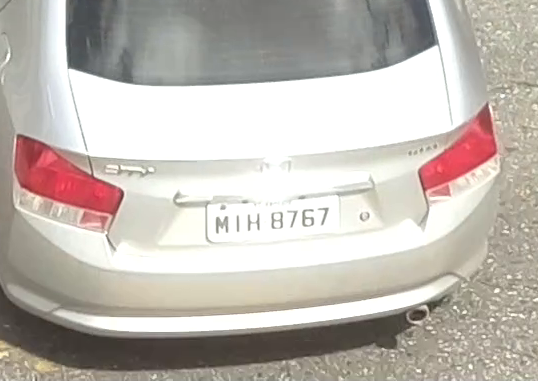}
};

\draw(4.5,2.4) node[text centered, text width=2.2cm, inner sep=0pt] (cruz2-img6) {
\includegraphics[width=1.3cm,height=1.3cm]{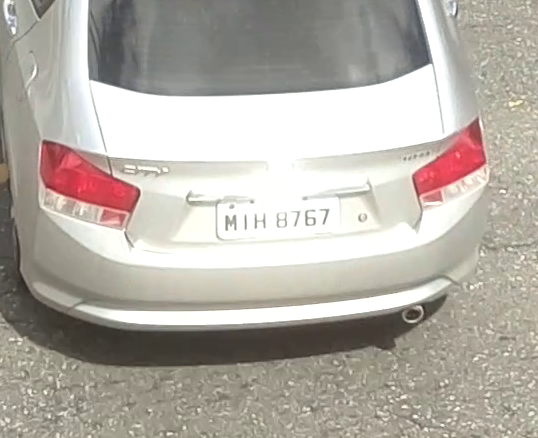}
};

\draw (cruz1-img1.north) -- (cruz2-img1.south);
\draw (cruz1-img1.north) -- (cruz2-img2.south);
\draw (cruz1-img1.north) -- (cruz2-img6.south);

\draw (cruz1-img2.north) -- (cruz2-img1.south);
\draw (cruz1-img2.north) -- (cruz2-img2.south);
\draw (cruz1-img2.north) -- (cruz2-img6.south);

\draw (cruz1-img6.north) -- (cruz2-img1.south);
\draw (cruz1-img6.north) -- (cruz2-img2.south);
\draw (cruz1-img6.north) -- (cruz2-img6.south);

\draw [decoration={brace},decorate] (-0.8,3.4) -- (5.3,3.4) node [pos=0.5,anchor=south,yshift=0.05cm] {$N$ images};

\path[->](3.1,2.4) node[] {\Large $\dots$};
\path[->](3.1,0.0) node[] {\Large $\dots$};
\path[->](-1.0,2.4) node[rotate=90] {Camera 1};
\path[->](-1.0,0.0) node[rotate=90] {Camera 2};

\path[->](0.0,-0.8) node[] {\small frame $j$};
\path[->](1.6,-0.8) node[] {\small frame $j$+1};
\path[->](4.5,-0.8) node[] {\small frame $j$+$N$};

\path[->](0.0,3.2) node[] {\small frame $i$};
\path[->](1.6,3.2) node[] {\small frame $i$+1};
\path[->](4.5,3.2) node[] {\small frame $i$+$N$};

\node[draw,align=left] at (6.4,1.3) { \small
All $N^2$ pairs\\
\small of each  \\
\small vehicle are\\
\small used to train\\
\small or test\\
\small (not for both)};

\end{tikzpicture}
\caption{Pair-wise data augmentation only for positive (true) matchings: from $N$ images of two vehicle sequences we extract $N^2$ distinct pairs. The same procedure is applied for license plate and vehicle shapes.}
\label{fig:pairwise}
\end{figure}

There are multiple distinct occurrences of the same vehicle as it moves across the video. Therefore, instead of only 869 matchings as shown in Table~\ref{tab:dataset}, we can generate thousands of true matchings by doing the Cartesian product between a sequence of images of the same vehicle that appears in Camera 1 and 2. This data augmentation is usually necessary for CNN training. Therefore, we used the MOSSE tracker~\cite{5539960} to extract the $N$ first  occurrences of each license plate (see Fig~\ref{fig:pairwise}).
Note that negative pairs are easier to generate, since we can use any combination of distinct vehicles from Camera 1 and 2.

We also adjusted another parameter $\lambda$ that was meant to multiply the number of false negatives pairs (non-matchings) in the testing set to simulate the network in a real environment assuming it may have many more tests of non-matchings pairs than the opposite. In Table~\ref{tab:settings} we show some parameter settings for our experiments. Note however, that we keep the same proportion of positive and negative pairs during the training in order to avoid class imbalance.
\begin{table}[!htpb]
   \setlength{\tabcolsep}{0.2pt}
   \def\mc{\multicolumn{2}{c||}}
   \def\df{\normalsize}
   \def\de{\normalsize}

   \centering
   \caption{Parameter settings used in our experiments.}
   \begin{tabular}{||l||c|c||c|c||c|r||}
         \hline
           \de Settings    &  \mc{Training} & \mc{Testing}  \\ \hline
                  &  \df \#positives & \df \#negatives & \df \#positives & \df \#negatives  \\ \hline
          \de $N = 3, \lambda = 5$ & \de 3867 & \de  3867 & \de 3903 &  \de 19515 \\
         \de $N = 10, \lambda = 10$ & \de 42130 & \de  42130 & \de 42707 &  \de 427070 \\
        \hline
   \end{tabular}
   \label{tab:settings}
 \end{table}

\begin{figure}[!htb]
\begin{tikzpicture}


\node[draw,align=left,text width=8.5cm,text height=0.2cm, inner sep=1pt] at (3.3,1.7) {
\small Siamese-Car (Stream 1): \textbf{matching} {\normalsize \cmark}\\
\small Siamese-Plate (Stream 2): \textbf{matching} {\normalsize \cmark}\\
\small Siamese (Two-Stream): \textbf{matching} {\normalsize \cmark}  };

\draw(0.0,3.4) node[text centered, text width=2.0cm] (img1) {
\includegraphics[width=2.1cm,height=2.1cm]{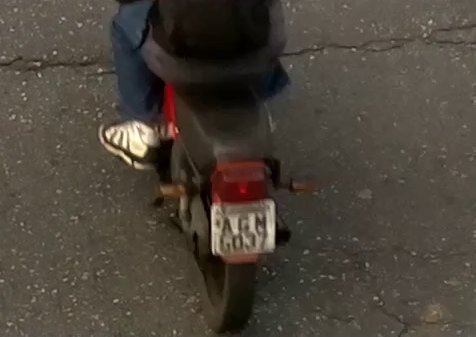}
};

\draw(2.2,3.4) node[text centered, text width=2.2cm] (img2) {
\includegraphics[width=1.6cm,height=1.5cm]{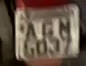}
};

\draw(4.35,3.4) node[text centered, text width=2.2cm] (img3) {
\includegraphics[width=2.1cm,height=2.1cm]{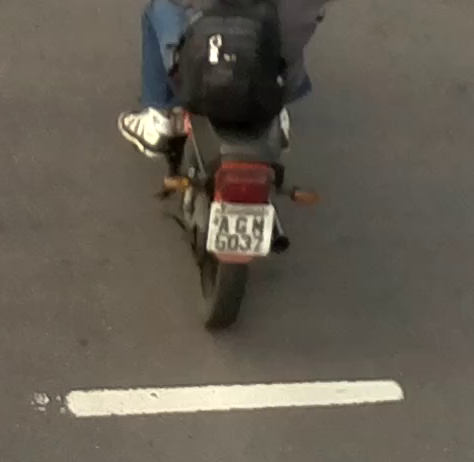}
};

\draw(6.5,3.4) node[text centered, text width=2.2cm] (img4) {
\includegraphics[width=1.6cm,height=1.5cm]{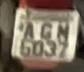}
};


\node[draw,align=left,text width=8.5cm,text height=0.2cm, inner sep=1pt] at (3.3,5.1) {
\small Siamese-Car (Stream 1): \textbf{non-matching} { \normalsize \cmark} \\
\small Siamese-Plate (Stream 2): \textbf{matching} { \normalsize \xmark} \\
\small Siamese (Two-Stream): \textbf{non-matching} { \normalsize \cmark}  };

\draw(0.0,6.8) node[text centered, text width=2.0cm] (img1) {
\includegraphics[width=2.1cm,height=2.1cm]{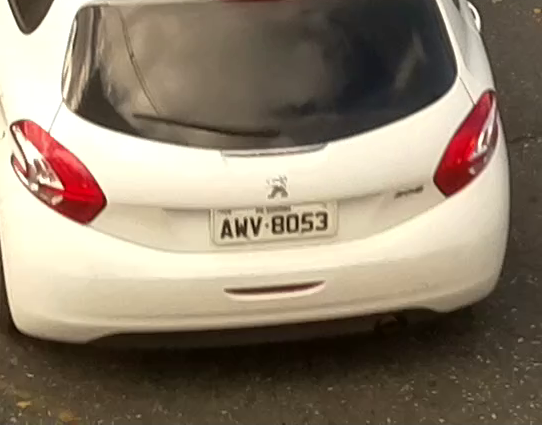}
};

\draw(2.2,6.8) node[text centered, text width=2.2cm] (img2) {
\includegraphics[width=2.1cm,height=1.0cm]{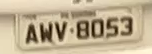}
};

\draw(4.35,6.8) node[text centered, text width=2.2cm] (img3) {
\includegraphics[width=2.1cm,height=2.1cm]{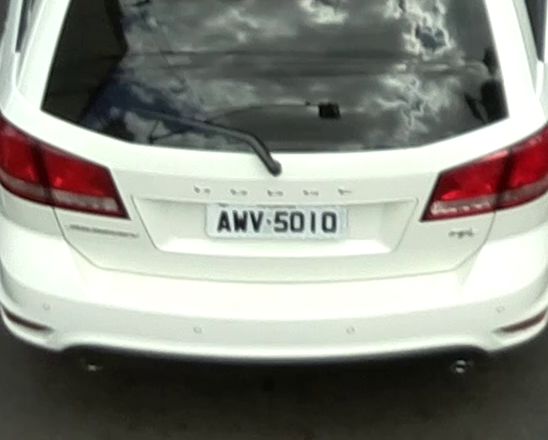}
};

\draw(6.5,6.8) node[text centered, text width=2.2cm] (img4) {
\includegraphics[width=2.1cm,height=1.0cm]{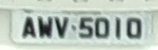}
};


\node[draw,align=left,text width=8.5cm,text height=0.2cm, inner sep=1pt] at (3.3,8.5) {
\small Siamese-Car (Stream 1): \textbf{matching} { \normalsize \xmark}\\
\small Siamese-Plate (Stream 2): \textbf{non-matching} { \normalsize \cmark}\\
\small Siamese (Two-Stream): \textbf{non-matching} { \normalsize \cmark} };

\draw(0.0,10.2) node[text centered, text width=2.0cm] (img1) {
\includegraphics[width=2.1cm,height=2.1cm]{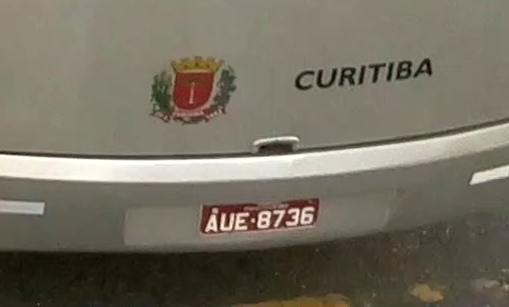}
};

\draw(2.2,10.2) node[text centered, text width=2.2cm] (img2) {
\includegraphics[width=2.1cm,height=1.0cm]{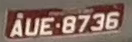}
};

\draw(4.35,10.2) node[text centered, text width=2.2cm] (img3) {
\includegraphics[width=2.1cm,height=2.1cm]{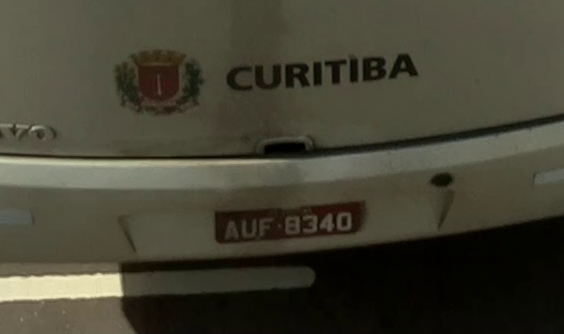}
};

\draw(6.5,10.2) node[text centered, text width=2.2cm] (img4) {
\includegraphics[width=2.1cm,height=1.0cm]{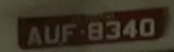}
};


\node[draw,align=left,text width=8.5cm,text height=0.2cm, inner sep=1pt] at (3.3,11.9) {
\small Siamese-Car (Stream 1): \textbf{non-matching} { \normalsize \xmark} \\
\small Siamese-Plate (Stream 2): \textbf{non-matching} { \normalsize \xmark}\\
\small Siamese (Two-Stream): \textbf{non-matching} { \normalsize \xmark}};

\draw(0.0,13.6) node[text centered, text width=2.0cm] (img1) {
\includegraphics[width=2.1cm,height=2.1cm]{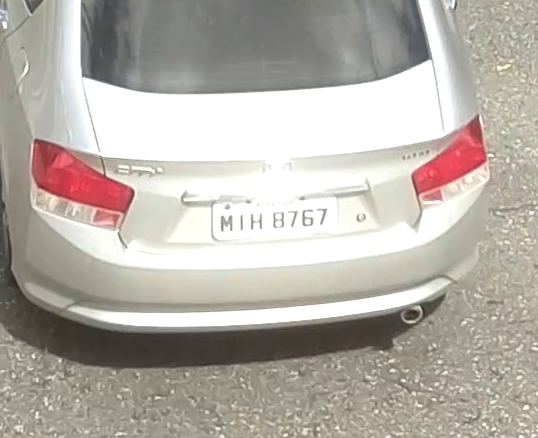}
};

\draw(2.2,13.6) node[text centered, text width=2.2cm] (img2) {
\includegraphics[width=2.1cm,height=1.0cm]{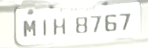}
};

\draw(4.35,13.6) node[text centered, text width=2.2cm] (img3) {
\includegraphics[width=2.1cm,height=2.1cm]{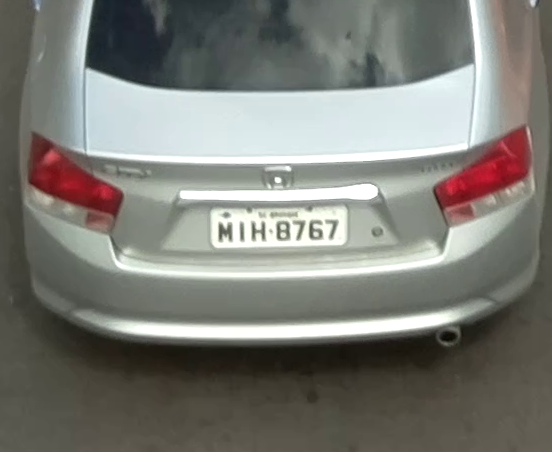}
};

\draw(6.5,13.6) node[text centered, text width=2.2cm] (img4) {
\includegraphics[width=2.1cm,height=1.0cm]{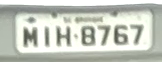}
};

\end{tikzpicture}
\caption{Inference results (testing set): from top-to-bottom an example where the three architectures failed (severe lighting conditions); Siamese-Car failed (similar vehicle shape); Siamese-Plate failed (similar license plate); and, at bottom, the three architectures found a correct matching.}
\label{fig:results}
\end{figure}

The quantitative criteria we used to evaluate the architectures performance are the precision $P$, recall $R$, accuracy $A$
and $F$-measure. As shown in Table~\ref{tb:results}, the Two-Stream Siamese outperforms two distinct One-Stream Siamese Networks: the first one, Siamese-Car, is fed only with the shape of vehicles ($96 \times 96$ pixels); the second, Siamese-Plate, only use patches of license plates ($96 \times 48$ pixels). Note that even when we increased the number of false matchings in the negative testing set, $\lambda=10$, the $F$-measure of the Two-Stream Siamese was similar in both scenarios. The accuracy $A$ is usually much higher since the number of negative pairs is much larger. Some inference results are shown in Fig.~\ref{fig:results}.



\begin{table}[ht]

\setlength{\tabcolsep}{4pt}
\def\mc{\multicolumn{4}{c|}}
\def\de{\small}
\centering
\caption{Matching performance of the proposed Two-Stream Siamese (Small-VGG) against two One-Stream Siamese (Car and Plate with Small-VGG) by using different settings to generate image pairs.}
\label{tb:results}
\begin{tabular}{|l||c|c|c|c|} \hline
& \mc{$N = 3, \lambda = 5$}  \\ \hline
\de Algorithm  & \de   $P$   &\de   $R$   &\de   $F$   &\de   $A$ \\ \hline
\de Siamese-Car \hspace{3pt} (Stream 1) &\de   85.8\%   &\de   93.1\%   &\de   89.3\%   &\de    96.3\%  \\ \hline
\de Siamese-Plate (Stream 2) &\de   75.9\%   &\de   81.8\%   &\de   78.8\%   &\de    92.6\%  \\ \hline
\de Siamese \hspace{18pt} (Two-Stream) &\de   92.7\%   &\de   93.0\%   &\de   92.9\%   &\de    97.6\%  \\ \hline
& \mc{$N = 10, \lambda = 10$}  \\ \hline
\de Algorithm  & \de   $P$   &\de   $R$   &\de   $F$   &\de   $A$ \\ \hline
\de Siamese-Car \hspace{3pt} (Stream 1) &\de   92.4\%   &\de   83.5\%   &\de   87.8\%   &\de    97.9\%  \\ \hline
\de Siamese-Plate (Stream 2) &\de   86.8\%   &\de   59.5\%   &\de   70.6\%   &\de    95.5\%  \\ \hline
\de Siamese \hspace{18pt} (Two-Stream) &\de   94.7\%   &\de   90.6\%   &\de   92.6\%   &\de    98.7\%  \\ \hline
\end{tabular}
\end{table}

\begin{figure}[!htb]
\hspace{-10pt}
\begin{tikzpicture}

\draw(-0.4,2.53) node[text centered, text width=3.2cm] (img1) {
{\small Vehicle (96$\times$96 pixels)}\\
\vspace{0.1cm}
\includegraphics[width=2.1cm,height=2.1cm]{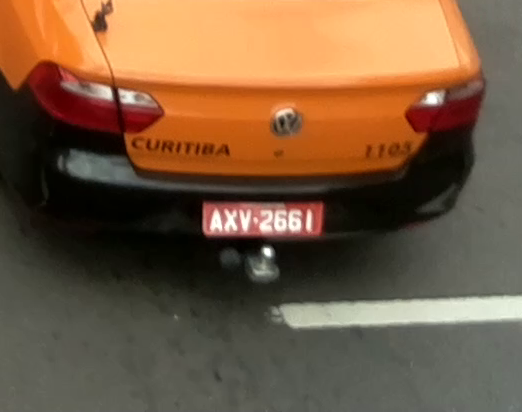}
};

\draw(-0.4,0.3) node[text centered, text width=3.2cm] (img2) {
{\small Plate (96$\times$48 pixels)}\\
\vspace{0.1cm}
\includegraphics[width=2.1cm,height=1.0cm]{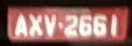}
};

\draw(4.35,1.5) node[text centered, text width=4.5cm] (img3) {
{\small Vehicle (patches 224$\times$224 pixels)}\\
\vspace{0.1cm}
\includegraphics[width=4.2cm,height=4.2cm]{./figures/two-stream/fig4_8336c2.png}
};

\draw(-0.2,-1.3) node[text centered, text width=4.0cm] (t1) {
{\small Siamese Two-Stream\\ (\textbf{Small-VGG})\\  $F$ = 92.6\% and $A$ 98.7\%}};\\
\vspace{0.1cm}
\draw(4.35,-1.2) node[text centered, text width=4.4cm] (t2) {
{\small Siamese-Car (\textbf{Small-VGG}):  $F$ = 88.1\% and $A$ = 97.9\%}};\\
\vspace{0.1cm}

\draw(1.5,1.4) node[text centered, text width=4.4cm] (vs) {
\includegraphics[scale=0.6]{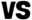}
};
\draw(4.35,-2.0) node[text centered, text width=4.4cm] (t3) {
{\small Siamese-Car (\textbf{Resnet50}):\\  $F$ = 81.2\% and $A$ = 97.1\%}};
\end{tikzpicture}
\caption{Siamese Two-Stream versus Siamese-Car.}
\label{fig:results2}
\end{figure}

We also tried different CNN in our Two-Stream Siamese, their performance are reported in Table~\ref{tb:results2}.
Furthermore, as can be seen in Fig.~\ref{fig:results2}, we also evaluated the performance of the proposed Two-Stream Siamese against two One-Stream Siamese versions fed with larger image patches ($224 \times 224$ pixels). Note that we achieved a higher $F$-measure by using two small image patches than a single patch containing both features. Another advantage is the Two-Stream Siamese training time: 1938 seconds per epoch ($N = 10$ and $\lambda = 10$) against 3441 seconds per epoch of the Siamese-Car by using the same Small-VGG and 4937 seconds with ResNet. 
The experiments were carried out on a Intel i7 with 32GB DRAM and a Nvidia Titan Xp GPU.


\begin{table}[ht]

\setlength{\tabcolsep}{5pt}
\def\mc{\multicolumn{4}{c|}}
\def\de{\small}
\centering
\caption{Matching performance of the proposed Two-Stream Siamese with different CNN architectures.}
\label{tb:results2}
\begin{tabular}{|l||c|c|c|c|} \hline
    & \mc{$N = 10, \lambda = 10$}  \\ \hline
\de Siamese (Two-Stream)  & \de   $P$   &\de   $R$   &\de   $F$   &\de   $A$ \\ \hline
\de CNN = Lenet5 \hspace{3pt} &\de   89.6\%   &\de   85.2\%   &\de  87.3\%   &\de  97.8\%  \\ \hline
\de CNN = Matchnet~\cite{matchnet2015} \hspace{3pt}  &\de   94.5\%   &\de   87.1\%   &\de   90.7\%   &\de    98.4\%   \\ \hline
\de CNN = MC-CNN~\cite{mccnn} \hspace{3pt}  &\de   89.0\%   &\de   90.1\%   &\de   89.6\%   &\de    98.1\%   \\ \hline
\de CNN = GoogleNet \hspace{3pt}  &\de   88.8\%   &\de   81.8\%   &\de   85.1\%   &\de    97.4\%   \\ \hline
\de CNN = AlexNet \hspace{3pt}  &\de   91.3\%   &\de   86.5\%   &\de   88.8\%   &\de    98.0\%   \\ \hline
\de CNN = Small-VGG \hspace{3pt}  &\de   94.7\%   &\de   90.6\%   &\de   92.6\%   &\de    98.7\%   \\ \hline
\end{tabular}
\end{table}

\section{Conclusions}~\label{sec:conclusions}

We proposed in this paper a fast Two-Stream Siamese that combines the discriminatory power of two distinctive and persistent features, the vehicle's shape and the registration plate, to address the problem of vehicle re-identification by using non-overlapping cameras. Tests indicate that our network is more robust than other One-Stream Siamese architectures which are fed with the same features or larger images. We also evaluated simple and complex CNNs, used by the Siamese Network, to find a trade-off between efficiency and performance.

\paragraph*{Acknowledgment:}
We gratefully acknowledge the support of NVIDIA Corporation with the donation of the Titan Xp GPU and the agencies CNPq, CAPES and SETRAN-Curitiba.

\bibliographystyle{IEEEbib}
\bibliography {article}
\end{document}